\documentclass{article}

\usepackage{times}
\usepackage{graphicx} 
\usepackage{subfigure} 

\usepackage{natbib}

\usepackage{algorithm}
\usepackage{algorithmic}

\usepackage{hyperref}
\usepackage{url}
\usepackage{graphicx}
\usepackage{amsmath,amsfonts,amssymb}
\usepackage{mathtools}
\usepackage{amsthm}
\usepackage{color}

\usepackage[accepted]{icml2016}

\newcommand{\matr}[1]{\mathbf{#1}}
\newcommand\RR{\mathbb{R}}
\newcommand\CC{\mathbb{C}}
\newtheorem{lemma}{Lemma}
\newcommand\norm[1]{\left\lVert#1\right\rVert}

\newcommand*\samethanks[1][\value{footnote}]{\footnotemark[#1]}

\begin{document} 

\twocolumn[
\icmltitle{Unitary Evolution Recurrent Neural Networks}

\icmlauthor{Martin Arjovsky \thanks{}}{marjovsky@dc.uba.ar}
\icmlauthor{Amar Shah \samethanks}{as793@cam.ac.uk}

\icmlauthor{Yoshua Bengio}{}
\icmladdress{Universidad de Buenos Aires, University of Cambridge, \\
Universit\'e de Montr\'eal. Yoshua Bengio is a CIFAR Senior Fellow.}

\samethanks Indicates first authors. Ordering determined by coin flip.

\vskip 0.3in
]

\begin{abstract} 
Recurrent neural networks (RNNs) are notoriously difficult to train. When the eigenvalues of the hidden to hidden weight matrix
deviate from absolute value 1, optimization becomes difficult due to the well studied issue of {\it{vanishing}} and {\it{exploding}} gradients, especially when trying to learn long-term dependencies.
To circumvent this problem, we propose a new architecture that learns a unitary weight matrix, with eigenvalues
of absolute value exactly 1. The challenge we address is that of parametrizing unitary matrices in a way that does not require expensive computations (such as eigendecomposition) after each weight update. We construct an expressive unitary weight matrix by composing several structured matrices that act
as building blocks with parameters to be learned. Optimization with this parameterization becomes feasible only when considering hidden
states in the complex domain. We demonstrate the potential of this architecture by achieving state of the art
results in several hard tasks
involving very long-term dependencies.
\end{abstract} 

\section{Introduction}
Deep Neural Networks have shown remarkably good performance on a wide range of complex data problems 
including speech recognition \citep{Hinton2012}, image recognition \citep{Krizhevsky2012} and natural 
language processing \citep{Collobert2011}. However, training very deep models remains a difficult task. 
The main issue surrounding the 
training of deep networks is the {\it{vanishing}} and {\it{exploding}} gradients problems
introduced by~\citet{Hochreiter91-small} and shown by~\citet{Yoshua94} to be necessarily arising when trying to learn
to reliably store bits of information in any parametrized dynamical system.
If gradients propagated back through a network vanish, the credit assignment role of backpropagation
is lost, as information about small changes in states in the far past has no influence on future states.
If gradients explode, gradient-based optimization algorithms struggle to 
traverse down a cost surface, because gradient-based optimization assumes small changes in parameters
yield small changes in the objective function.
As the number of time steps
considered in the sequence of states grows, the shrinking or expanding effects associated with the state-to-state
transformation at individual time steps can grow exponentially, yielding respectively vanishing or exploding
gradients. See~\citet{Pascanu2013} for a review.

Although the long-term dependencies problem appears intractable in the absolute~\citep{Yoshua94} for
parametrized dynamical systems, several heuristics have recently been found to help 
reduce its effect, such as the use of self-loops and gating units in the LSTM~\citep{LSTM}
and GRU~\citep{Cho2014a} recurrent architectures.
Recent work also supports the idea of using \textit{orthogonal} weight matrices to assist optimization  
\citep{Saxe2014, Quoc2015}.

In this paper, we explore the use of orthogonal and unitary matrices in recurrent neural networks.
We start in Section~\ref{sec:ortho} by showing a novel bound on the propagated gradients
in recurrent nets when the recurrent matrix is orthogonal.
Section~\ref{sec:uRNN} discusses the difficulties of parameterizing real valued orthogonal matrices and how
they can be alleviated by moving to the \textit{complex} domain. 

We discuss a novel approach to constructing expressive unitary matrices as the composition of simple
unitary matrices which require at most $\mathcal{O}(n \log n)$ computation and $\mathcal{O}(n)$ memory,
when the state vector has dimension $n$. These are
unlike general matrices, which require $\mathcal{O}(n^2)$ computation and memory. 
Complex valued representations have been considered for neural networks in the past, 
but with limited success and adoption \citep{hirose2003complex, zimmermann2011comparison}. 
We hope our findings will change this.

Whilst our model uses complex valued matrices and parameters, all
implementation and optimization is possible with real numbers and has been
done in Theano \citep{Fred2010}. This along with other implementation details are discussed
in Section~\ref{sec:impl}, and the code used for the experiments is available online.
The potential of the developed model for learning long term dependencies
with relatively few parameters is explored in Section~\ref{expts}. 
We find that the proposed architecture generally outperforms LSTMs and previous
approaches based on orthogonal initialization.

\section{Orthogonal Weights and Bounding the Long-Term Gradient}
\label{sec:ortho}

A matrix, $\matr{W}$, is orthogonal if 
$\matr{W}^\top \matr{W} = \matr{W} \matr{W}^\top = \matr{I}$. 
Orthogonal matrices have the property that they preserve norm (i.e. $\| \matr{W} h \|_2 = \| h \|_2$)
and hence repeated iterative multiplication of a vector by an orthogonal matrix leaves the norm of the 
vector unchanged.

Let $h_T$ and $h_t$ be the hidden unit vectors for hidden layers $T$ and $t$ of a neural network with 
$T$ hidden layers and $T \gg t$. 
If $C$ is the objective we are trying to minimize, then the {\it{vanishing}} and {\it{exploding}} 
gradient problems refer to the decay or growth of $\frac{\partial C}{\partial h_t}$ as the 
number of layers, $T$, grows. Let $\sigma$ be a pointwise nonlinearity function, and
\begin{align}
  z_{t+1} &= \matr{W}_t h_t + \matr{V}_t x_{t+1}  \notag \\
  h_{t+1} &= \sigma (z_{t+1})
\label{nonlinoutput}
\end{align}
then by the chain rule
\begin{align}
  \frac{\partial C}{\partial h_t} &= \frac{\partial C}{\partial h_T} \frac{\partial h_T}{\partial h_t} \notag \\
  &= \frac{\partial C}{\partial h_T} \prod_{k=t}^{T-1} \frac{\partial h_{k+1}}{\partial h_k} 
    = \frac{\partial C}{\partial h_T} \prod_{k=t}^{T-1} \matr{D}_{k+1} \matr{W}_k^T 
\end{align}
where $\matr{D}_{k+1} = diag(\sigma'(z_{k+1}))$ is the Jacobian matrix of the pointwise nonlinearity.

In the following we define the norm of a matrix to refer to the spectral radius norm (or operator 2-norm)
and the norm of a vector to mean $L_2$-norm. By definition of the operator norms, 
for any matrices $\matr{A}, \matr{B}$ and vector $v$ we have $\norm{\matr{A}v} \leq \norm{\matr{A}} \norm{v}$ and $\norm{\matr{A}\matr{B}} \leq \norm{\matr{A}} \norm{\matr{B}}$.
If the weight matrices $\matr{W}_k$ are norm preserving (i.e. orthogonal), then we {\bf{prove}}
\begin{align}
  \norm{ \frac{\partial C}{\partial h_t} } &= \norm{ \frac{\partial C}{\partial h_T} 
    \prod_{k=t}^{T-1} \matr{D}_{k+1} \matr{W}_k^T } \notag \\ 
  &\leq \norm{\frac{\partial C}{\partial h_T}} 
    \prod_{k=t}^{T-1} \norm{ \matr{D}_{k+1} \matr{W}_k^T } \notag \\
  &= \norm{ \frac{\partial C}{\partial h_T}} \prod_{k=t}^{T-1} \norm{\matr{D}_{k+1}}. 
\label{bound}
\end{align}

Since $\matr{D}_k$ is diagonal, $\norm{\matr{D}_k} = \max_{j=1, ..., n} |\sigma'(z_k^{(j)})|$,
with $z_k^{(j)}$ the $j$-th pre-activation of the $k$-th hidden layer.
If the absolute value of the derivative $\sigma'$ can take some value $\tau > 1$, then 
this bound is useless, since $\|\frac{\partial C}{\partial h_t}\| 
\leq \norm{\frac{\partial C}{\partial h_T}} \tau^{T-t}$ which grows exponentially 
in $T$. We therefore cannot effectively bound $\frac{\partial C}{\partial h_t}$ 
for deep networks, resulting potentially in exploding gradients.

In the case $|\sigma'| < \tau < 1$, equation \ref{bound} proves that 
that $\frac{\partial C}{\partial h_t}$ tends to 0 exponentially fast as $T$ grows, 
resulting in guaranteed vanishing gradients. 
This argument makes the rectified linear unit (ReLU) nonlinearity an attractive choice
\citep{Glorot2011, Nair2010}. Unless all the activations are killed at one layer, 
the maximum entry of $\matr{D}_k$ is 1, resulting in
$\norm{\matr{D}_k} = 1$ for all layers $k$. With ReLU nonlinearities, we thus have
\begin{equation}
  \norm{\frac{\partial C}{\partial h_t}} \leq \norm{ \frac{\partial C}{\partial h_T}} 
  \prod_{k=t}^{T-1} \norm{\matr{D}_{k+1}} = \norm{\frac{\partial C}{\partial h_T}}.
\label{bound2}
\end{equation}

Most notably, this result holds for a network of arbitrary depth and renders engineering tricks
like gradient clipping unnecessary \citep{Pascanu2013}. 

To the best of our knowledge, this analysis is a novel contribution and the first time a 
neural network architecture has been mathematically proven to avoid exploding gradients. 

\vspace{-1mm}

\section{Unitary Evolution RNNs}
\label{sec:uRNN}
\vspace{-1mm}
Unitary matrices generalize orthogonal matrices to the complex domain.
A complex valued, norm preserving matrix,
$\matr{U}$, is called a \textit{unitary} matrix and is such that 
$\matr{U}^* \matr{U} = \matr{U} \matr{U}^* = \matr{I}$, where $\matr{U}^*$ is the conjugate transpose
of $\matr{U}$.  Directly parametrizing the set of unitary matrices in such a way that gradient-based
optimization can be applied is not straightforward because a gradient step will typically yield
a matrix that is not unitary, and projecting on the set of unitary matrices (e.g., by performing
an eigendecomposition) generally costs $\mathcal{O}(n^3)$ computation when $\matr{U}$ is $n \times n$.

The most important feature of unitary and orthogonal matrices for our purpose is that they have eigenvalues
$\lambda_j$ with absolute value 1. The following lemma, proved in \cite{linalgbook}, may shed light on a 
method which can be used to efficiently span a large set of unitary matrices.

\begin{lemma}
  A complex square matrix $\matr{W}$ is unitary if and only if it has an eigendecomposition of the form
  $\matr{W} = \matr{V} \matr{D} \matr{V}^*$, where $^*$ denotes the conjugate transpose.
  Here, $\matr{V}, \matr{D} \in \mathbb{C}^{n \times n}$ are
  complex matrices, where $\matr{V}$ is unitary, and $\matr{D}$ is a diagonal such that $|\matr{D}_{j,j}|=1$. 
  Furthermore, $\matr{W}$ is a real orthogonal matrix if and only if for every eigenvalue $\matr{D}_{j,j} 
  = \lambda_j$ with eigenvector $v_j$, there is also a complex conjugate
  eigenvalue $\lambda_k = \overline{\lambda_j}$ with 
  corresponding eigenvector $v_k = \overline{v_j}$ .
\label{lemma}
\end{lemma}

Writing $\lambda_j = e^{i w_j}$ with $w_j \in \RR$, a naive method to learn a unitary matrix would be to
fix a basis of eigenvectors $\matr{V} \in \mathbb{C}^{n \times n}$ and set
\begin{equation} \matr{W} = \matr{V} \matr{D} \matr{V}^{*} , \end{equation}
\vspace{-0.5mm}
where $\matr{D}$ is a diagonal such that $\matr{D}_{j,j} = \lambda_j$. 

Lemma \ref{lemma} informs us how to construct a real orthogonal matrix, $\matr{W}$.
We must (i) ensure the columns of $\matr{V}$ come in complex conjugate pairs, $v_k = \overline{v_j}$, and
(ii) tie weights $w_k=-w_j$ in order to achieve $e^{i w_j} = \overline{e^{i w_k}}$. 
Most neural network objective functions are differentiable with respect to the weight matrices,
and consequently $w_j$ may be learned by gradient descent. 

Unfortunately the above approach has undesirable properties. 
Fixing $\matr{V}$ and learning $w$ requires $\mathcal{O}\left(n^2\right)$ memory, 
which is unacceptable given that the number of learned parameters is $\mathcal{O}(n)$. 
Further note that calculating $\matr{V} u$ for an arbitrary vector $u$ 
requires $\mathcal{O}(n^2)$ computation. 
Setting $\matr{V}$ to the identity would satisfy the conditions of the lemma, whilst reducing  
memory and computation requirements to $\mathcal{O}(n)$, however, $\matr{W}$ would remain diagonal, 
and have poor representation capacity.

We propose an alternative strategy to parameterize unitary matrices. 
Since the product of unitary matrices is itself a unitary matrix, we 
compose several simple, parameteric, unitary matrices to construct a single, expressive unitary matrix.
The four unitary building blocks considered are 

\begin{itemize}
  \item $\matr{D}$, a diagonal matrix with $\matr{D}_{j,j} = e^{i w_j}$, with parameters $w_j \in \RR$,
  \item $\matr{R} = \matr{I} - 2 \frac{v v^*}{\|v\|^2}$, a reflection matrix in the complex vector 
  $v \in \mathbb{C}^n$, 
  \item $\matr{\Pi}$, a fixed random index permutation matrix, and
  \item $\mathcal{F}$ and $\mathcal{F}^{-1}$, the Fourier and inverse Fourier transforms.
\end{itemize}

Appealingly, $\matr{D}$, $\matr{R}$ and $\matr{\Pi}$ all permit $\mathcal{O}(n)$ storage and 
$\mathcal{O}(n)$ computation for matrix vector products. $\mathcal{F}$ and $\mathcal{F}^{-1}$
require no storage and $\mathcal{O}(n \log n)$ matrix vector multiplication using the Fast Fourier
Transform algorithm. A major advantage of composing unitary matrices of the form listed above, is 
that the number of parameters, memory and computational cost increase almost linearly in the size
of the hidden layer. With such a weight matrix, immensely large hidden layers are feasible to train, 
whilst being impossible in traditional neural networks. 
 
With this in mind, in this work we choose to consider recurrent neural networks with unitary hidden to hidden
weight matrices. Our claim is that the ability to have large hidden layers where hidden 
states norms are preserved provides a powerful tool for modeling long term dependencies in sequence data. 
\cite{Yoshua94} suggest that having a large memory may be crucial for solving 
difficult tasks with long ranging dependencies: the smaller the state dimension, the more
information necessarily has to be eliminated when mapping a long sequence to a fixed-dimension state.

We call any RNN architecture which uses a unitary hidden to hidden matrix a \textit{unitary evolution RNN}
(uRNN). After experimenting with several structures, we settled on the following composition
\vspace{-0.25mm}
\begin{equation} 
\matr{W} = \matr{D}_3 \matr{R}_2 \mathcal{F}^{-1} \matr{D}_2 \matr{\Pi} \matr{R}_1 \mathcal{F} \matr{D}_1 .
\end{equation}
\vspace{-0.5mm}
Whilst each but the permutation matrix is complex, we parameterize and represent them with real numbers
for implementation purposes. When the final cost is real and differentiable, we may perform gradient descent 
optimization to learn the parameters.
\cite{dfc} construct a real valued, non-orthogonal matrix using a similar parameterization with the motivation
of parameter reduction by an order of magnitude on an industrial sized network. This combined with earlier
work \citep{fastfood} suggests that it is possible to create highly expressive matrices by composing simple
matrices with few parameters. In the following section, we explain details on how to implement our model 
and illustrate how we bypass the
potential difficulties of working in the complex domain.

\section{Architecture details}
\label{sec:impl}
\vspace{-0.1mm}
In this section, we describe the nonlinearity we used, how we incorporate real valued inputs 
with complex valued hidden units and map from complex hidden states to real outputs. 
\vspace{-0.1mm}
\subsection{Complex hidden units}
\vspace{-0.1mm}
Our implementation represents all complex numbers using real values in terms of their
real and imaginary parts. Under this framework, we sidestep the lack of support for complex numbers 
by most deep learning frameworks. Consider multiplying the complex weight matrix 
$\matr{W} = \matr{A} + i \matr{B}$ by the complex hidden vector $h = x + i y$, where
$\matr{A}, \matr{B}, x, y$ are real.
It is trivially true that $\matr{W}h = (\matr{A}x - \matr{B}y) + i (\matr{A}y + \matr{B}x)$.
When we represent $v \in \CC^n$ as $\big(Re(v)^\top, Im(v)^\top \big)^\top \in \RR^{2n}$ , we
compute complex matrix vector products with real numbers as follows
\vspace{-0.2mm}
\begin{equation} \begin{pmatrix} Re(\matr{W}h) \\ Im(\matr{W}h) \end{pmatrix}  
= \begin{pmatrix} \matr{A} & -\matr{B} \\ \matr{B} & \ \ \ 
\matr{A} \end{pmatrix} \begin{pmatrix} Re(h) \\ Im(h) \end{pmatrix} .
\end{equation}

More generally, let $f: \CC^n \rightarrow \CC^n$ be any complex function and $z = x + i y$ 
any complex vector. We may write $ f(z) = \alpha(x, y) + i \beta(x, y) $ where 
$\alpha, \beta : \RR^n \rightarrow \RR^n$. 
This allows us to implement everything using real valued operations, compatible with any
any deep learning framework with automatic differentiation such as Theano.

\subsection{Input to Hidden, Nonlinearity, Hidden to Output}

As is the case with most recurrent networks, our uRNN follows the same hidden to hidden mapping as 
equation \ref{nonlinoutput} with $\matr{V}_t = \matr{V}$ and $\matr{W}_t = \matr{W}$. 
Denote the size of the complex valued hidden states as $n_h$.
The input to hidden matrix is complex valued, $\matr{V} \in \CC^{n_h \times n_\mathrm{in}}$. 
We learn the initial hidden state $h_0 \in \CC^{n_h}$ as a parameter of the model.

Choosing an appropriate nonlinearity is not trivial in the complex domain.
As discussed in the introduction, using a ReLU is a natural choice in combination with a norm preserving
weight matrix. We first experimented with placing separate ReLU activations on the real and imaginary
parts of the hidden states.
However, we found that such a nonlinearity usually performed poorly.
Our intuition is that applying separate ReLU nonlinearities to the real 
and imaginary parts brutally impacts the 
phase of a complex number, making it difficult to learn structure.

We speculate that maintaining the phase of hidden states may be important for storing information 
across a large number of time steps, and our experiments supported this claim.
A variation of the ReLU that we name {\bf modReLU}, is what we 
finally chose. It is a pointwise nonlinearity,  
$\sigma_\mathrm{modReLU} (z) : \CC \rightarrow \CC$, which
affects only the absolute value of a complex number, defined as 
\begin{equation} \sigma_\mathrm{modReLU} (z) = 
\left\{
  \begin{array}{ll}
    (|z|+b) \frac{z}{|z|}  & \mbox{if } |z| + b \geq 0 \\
    0 & \mbox{if } |z| + b < 0
  \end{array}
\right.
\end{equation}

where $b \in \RR$ is a bias parameter of the nonlinearity. For a $n_h$ dimensional hidden space
we learn $n_h$ nonlinearity bias parameters, one per dimension. 
Note that the modReLU is similar to the ReLU in spirit, in fact more concretely
$\sigma_\mathrm{modReLU}(z) = \sigma_\mathrm{ReLU}(|z| + b) \frac{z}{|z|}$. 

To map hidden states to output, we define a matrix $\matr{U} \in \RR^{n_o \times 2n_h}$, 
where $n_o$ is the output dimension. We calculate a linear output as
\vspace{-0.5mm}
\begin{equation} o_t = \matr{U} \begin{pmatrix} Re(h_t) \\ Im(h_t) \end{pmatrix} + b_o , \end{equation}
\vspace{-0.25mm}
where $b_o \in \RR^{n_o}$ is the output bias. 
The linear output is real valued ($o_t \in \RR^{n_o}$) and can be used for prediction and loss function 
calculation akin to typical neural networks (e.g. it may be passed through a softmax which is used for 
cross entropy calculation for classification tasks).

\vspace{-1mm}
\subsection{Initialization}

Due to the stability of the norm preserving operations of our network, we found that performance was
not very sensitive to initialization of parameters.
For full disclosure and reproducibility, we explain our initialization strategy for each parameter below.
\vspace{-0.25mm}
\begin{itemize}
  \item We initialize $\matr{V}$ and $\matr{U}$ (the input and output matrices) as in \cite{Glorotinit},
  with weights sampled independently from uniforms, $\mathcal{U}\left[-\frac{\sqrt{6}}{\sqrt{n_{in}+ n_{out}}}, \frac{\sqrt{6}}{\sqrt{n_{in}+ n_{out}}}\right]$.
  \item The biases, $b$ and $b_o$ are initialized to 0. This implies that at initialization, 
    the network is linear with unitary weights, which seems to help early optimization \citep{Saxe2014}.
  \item The reflection vectors for $\matr{R}_1$ and $\matr{R}_2$ are initialized coordinate-wise from a 
  uniform $\mathcal{U}[-1, 1]$. Note that the reflection matrices are invariant to scalar multiplication 
  of the parameter vector, hence the width of the uniform initialization is unimportant.
  \item The diagonal weights for $\matr{D}_1, \matr{D}_2$ and $\matr{D}_3$ are sampled 
  from a uniform, $\mathcal{U}[-\pi, \pi]$. This ensures that the diagonal entries $\matr{D}_{j,j}$
  are sampled uniformly over the complex unit circle.
  \item We initialize $h_0$ with a uniform, 
  $\mathcal{U}\left[-\sqrt{\frac{3}{2n_h}}, \sqrt{\frac{3}{2n_h}} \right]$, 
  which results in $\mathbb{E}\left[\|h_0\|^2\right] = 1$. Since the norm of the hidden units are roughly 
  preserved through unitary evolution and inputs are typically whitened to have norm 1, 
  we have hidden states, inputs and linear outputs of the same order of magnitude, which seems to
  help optimization.
\end{itemize}

\section{Experiments}
\label{expts}

\begin{figure*}[t!]
  \centering
  \begin{minipage}[b]{0.5\linewidth}
    \centering
    \includegraphics[scale=0.25]{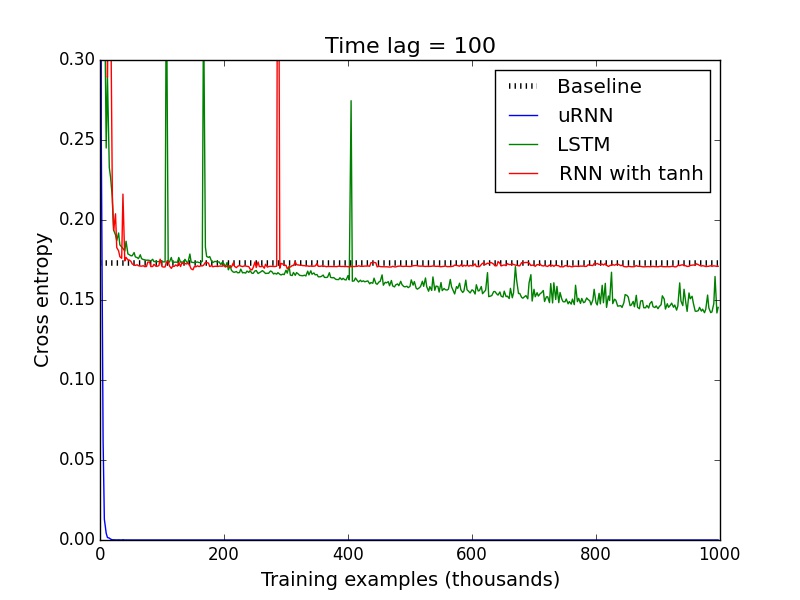}
  \end{minipage}
  \begin{minipage}[b]{0.5\linewidth}
    \centering
    \includegraphics[scale=0.25]{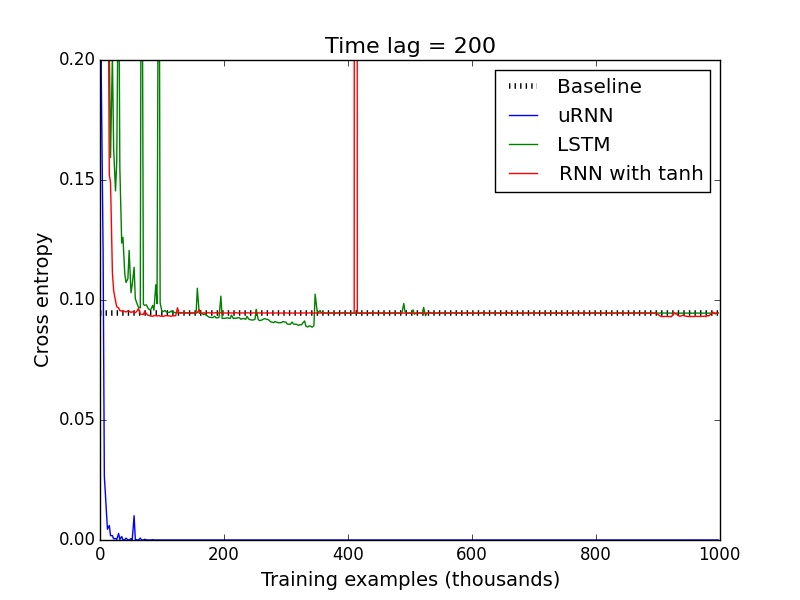}
  \end{minipage} 
  \begin{minipage}[b]{0.5\linewidth}
    \centering
    \includegraphics[scale=0.25]{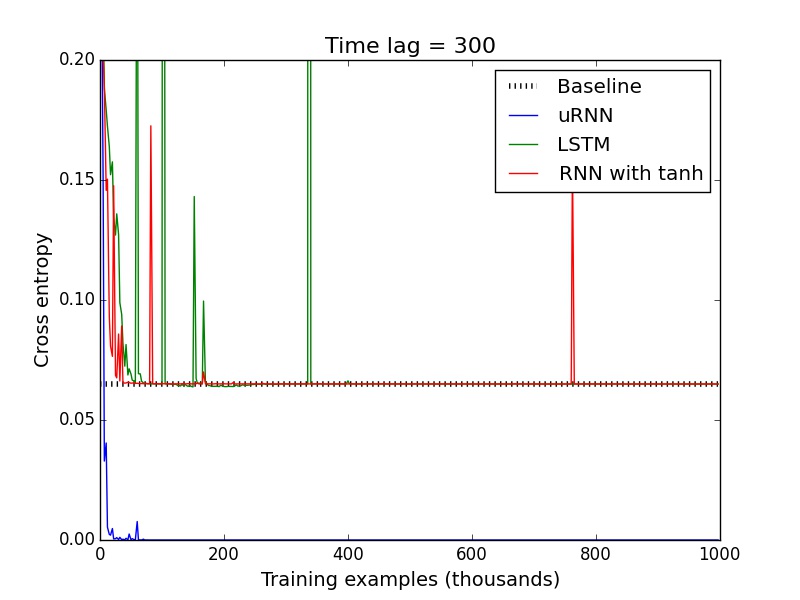}
  \end{minipage}
  \begin{minipage}[b]{0.5\linewidth}
    \centering
    \includegraphics[scale=0.25]{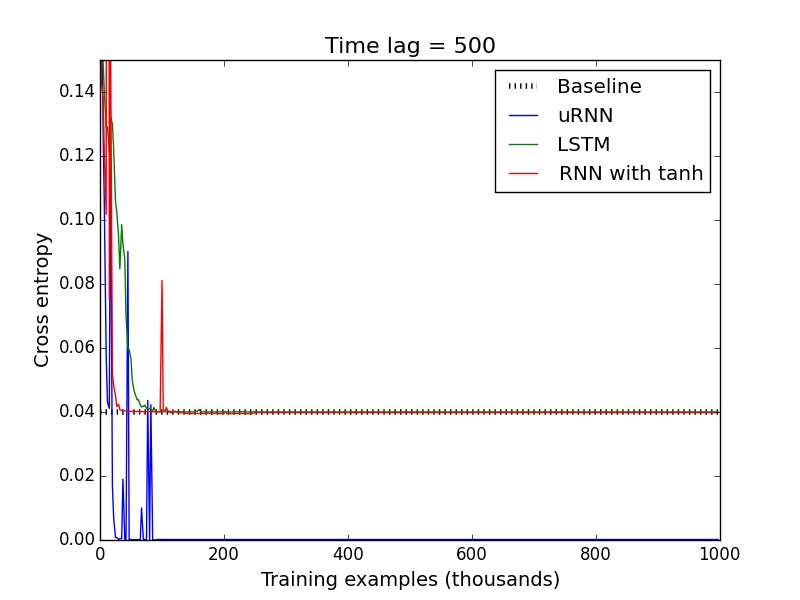}
  \end{minipage} 
  \caption{Results of the copying memory problem for time lags of $100, 200, 300, 500$. 
   The LSTM is able to beat the baseline only for $100$ times steps. Conversely the uRNN 
    is able to completely solve each time length in very few training iterations,
    without getting stuck at the baseline.}
  \label{fig1} 
\end{figure*}

In this section we explore the performance of our uRNN in relation to (a) RNN with tanh activations,
(b) IRNN \citep{Quoc2015}, that is 
an RNN with ReLU activations and with the recurrent weight matrix initialized to the identity, 
and (c) LSTM \citep{LSTM} models. We show that the uRNN shines quantitatively when it comes to modeling 
long term dependencies and exhibits qualitatively different learning properties to the other models. 
 
We chose a handful of tasks to evaluate the performance of the various models.
The tasks were especially created to be be pathologically hard, and have been used 
as benchmarks for testing the ability of a model to capture long-term memory \citep{LSTM, Quoc2015, NTM, HF}

Of the handful of optimization algorithms we tried on the various models, 
RMSProp \citep{RMSPROP} lead to fastest convergence and is what we stuck to for all
experiments here on in. However, we found the IRNN to be particularly unstable; it only ran without 
blowing up with incredibly low learning rates and gradient clipping. Since the performance was so poor
relative to other models we compare against, we do not show IRNN curves in the figures.  
In each experiment we use a learning rate of $10^{-3}$
and a decay rate of $0.9$. For the LSTM and RNN models, we had to clip gradients at 1 to avoid exploding 
gradients. Gradient clipping was unnecessary for the uRNN.

\subsection{Copying memory problem}
\begin{figure*}[t!] 
  \begin{minipage}[b]{0.5\linewidth}
    \centering
    \includegraphics[scale=0.25]{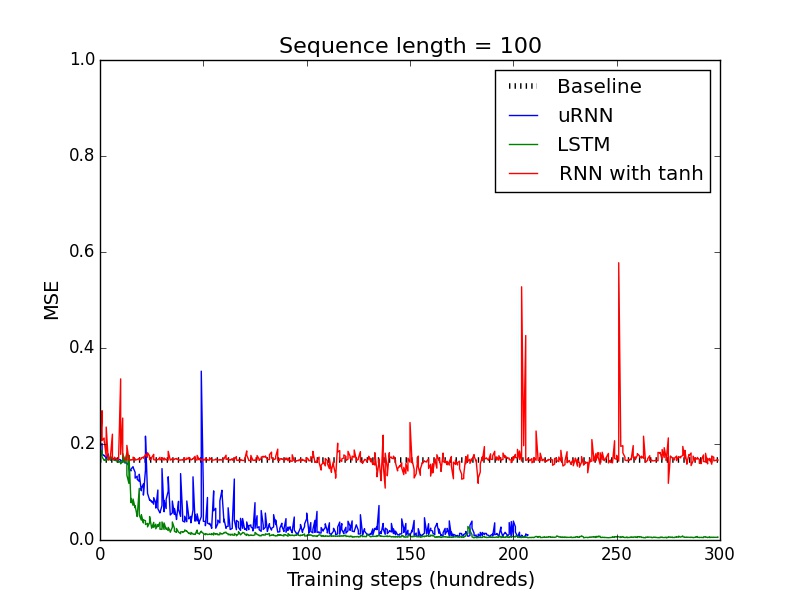}
  \end{minipage}
  \begin{minipage}[b]{0.5\linewidth}
    \centering
    \includegraphics[scale=0.25]{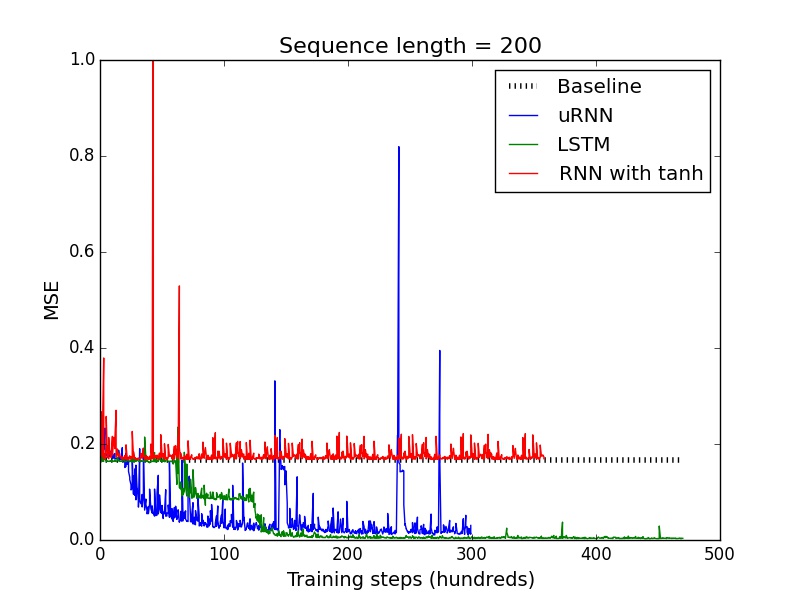}
    \end{minipage} 
  \begin{minipage}[b]{0.5\linewidth}
    \centering
    \includegraphics[scale=0.25]{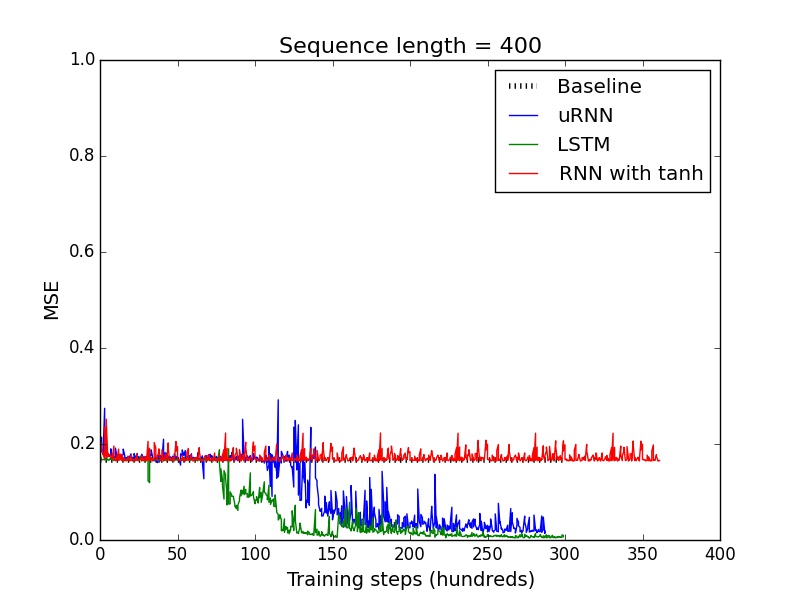}
    \end{minipage}
  \begin{minipage}[b]{0.5\linewidth}
    \centering
    \includegraphics[scale=0.25]{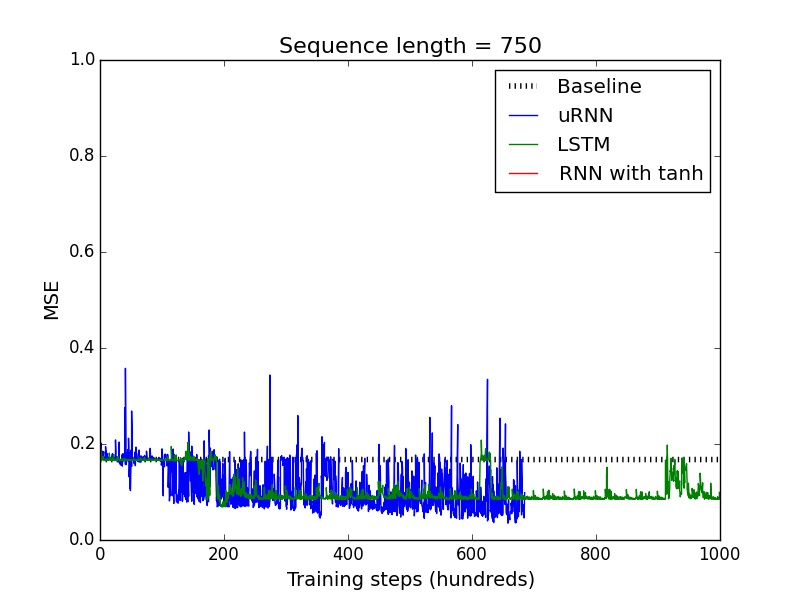}
  \end{minipage} 
  \caption{Results of the adding problem for $T=100, 200, 400, 750$. The RNN with tanh
	  is not able to beat the baseline for any time length. The LSTM and the uRNN show similar 
	  performance across time lengths, consistently beating the baseline.}
  \label{fig2} 
\end{figure*}

Recurrent networks have been known to have trouble remembering information about inputs seen
many time steps previously \citep{Yoshua94, Pascanu2013}. 
We therefore want to test the uRNN's ability to recall exactly data seen a long time ago.

Following a similar setup to \cite{LSTM}, we outline the copy memory task.
Consider 10 categories, $\{ a_i \}_{i=0}^9$.   
The input takes the form of a $T+20$ length vector of categories, where we test over a range of values
of $T$. 
The first $10$ entries are sampled uniformly, independently and with replacement from $\{a_i\}_{i=0}^7$,
and represent the sequence which will need to be remembered. 
The next $T-1$ entries are set to $a_8$, which can be thought of as the 'blank' category. 
The next single entry is $a_9$, which represents a delimiter, which should indicate to the algorithm
that it is now required to reproduce the initial $10$ categories in the output. 
The remaining $10$ entries are set to $a_8$. The required output sequence consists of $T+10$ 
repeated entries of $a_8$, followed by the first $10$ categories of the input sequence in exactly the
same order. The goal is to minimize the average cross entropy of category predictions 
at each time step of the sequence.
The task amounts to having to remember a categorical sequence of length 10, for $T$ time steps.

A simple baseline can be established by considering an optimal strategy when no memory is available, 
which we deem the \textit{memoryless} strategy. The memoryless strategy would be to predict $a_8$ for 
$T+10$ entries and then predict each of the final $10$ categories from the set $\{a_i\}_{i=0}^7$ independently
and uniformly at random. The categorical cross entropy of this strategy is $\frac{10 \log(8) }{T+20}$. 

We ran experiments where the RNN with tanh activations, IRNN, LSTM and uRNN had hidden layers of size 
80, 80, 40 and 128 respectively. This equates to roughly 6500 parameters per model. 
In Figure~\ref{fig1}, we see that aside from the simplest case, both the RNN with tanh and more surprisingly 
the LSTMs get almost exactly the same cost as the memoryless strategy. This behaviour is consistent 
with the results of \cite{NTM}, in which poor performance is reported for the LSTM for a very similar 
long term memory problem.

The uRNN consistently achieves perfect performance in relatively few iterations, even when having to recall
sequences after 500 time steps. What is remarkable is that the uRNN does not get stuck at the baseline at 
all, whilst the LSTM and RNN do. This behaviour suggests that the representations learned by the uRNN
have qualitatively different properties from both the LSTM and classical RNNs. 

\subsection{Adding Problem}

\begin{figure*}[t!]
  \begin{minipage}[b]{0.5\linewidth}
    \centering
    \includegraphics[scale=0.3]{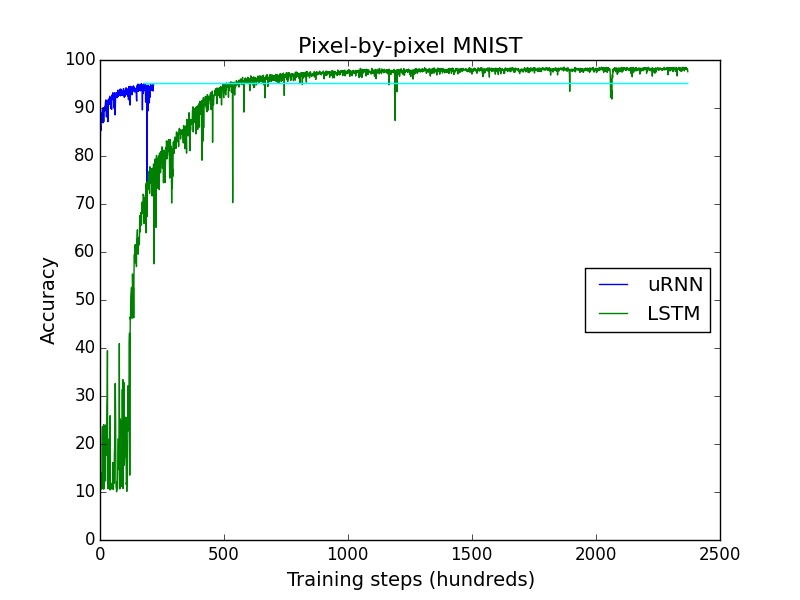}
  \end{minipage}
  \begin{minipage}[b]{0.5\linewidth}
    \centering
    \includegraphics[scale=0.3]{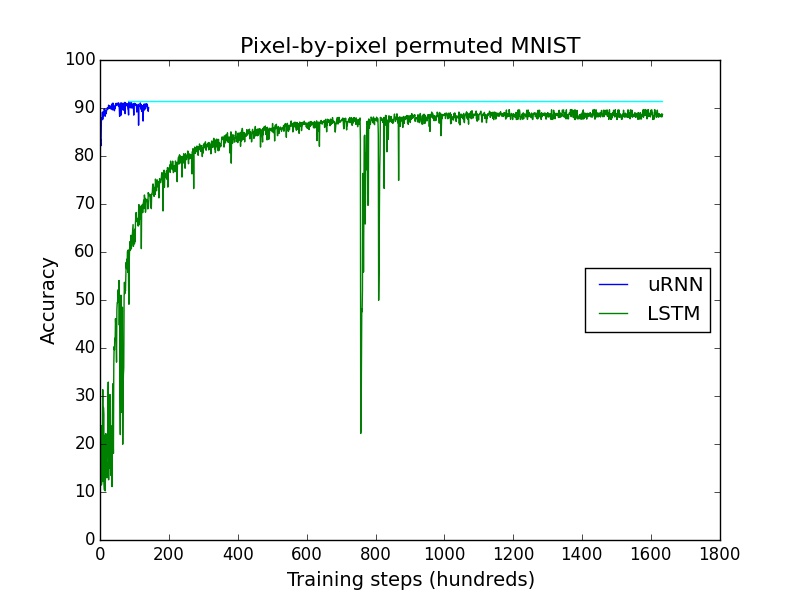}
  \end{minipage}
  \caption{Results on pixel by pixel MNIST classification tasks. 
           The uRNN is able to converge in a fraction of the iterations
	    that the LSTM requires. The LSTM performs better on MNIST classification, 
	    but the uRNN outperforms on the more complicated task
	    of permuted pixels.}
  \label{fig3} 
\end{figure*}

We closely follow the adding problem defined in \cite{LSTM} to explain the task at hand.
Each input consists of two sequences of length $T$. 
The first sequence, which we denote $x$, consists of numbers sampled uniformly at random 
$\mathcal{U}[0,1]$. 
The second sequence is an indicator sequence consisting of exactly two entries of 1 and remaining entries 0.
The first 1 entry is located uniformly at random in the first half of the sequence, whilst the second 1
entry is located uniformly at random in the second half.
The output is the sum of the two entries of the first sequence, corresponding to where the 1 entries are
located in the second sequence. A naive strategy of predicting 1 as the output regardless of the input sequence gives an expected mean
squared error of $0.167$, the variance of the sum of two independent uniform distributions.
This is our baseline to beat.

We chose to use 128 hidden units for the RNN with tanh, IRNN and LSTM and 512 for the uRNN.
This equates to roughly 16K parameters for the RNN with tanh and IRNN, 60K for the LSTM and almost 9K for
the uRNN. All models were trained using batch sizes of 20 and 50 with the best results being reported.
Our results are shown in Figure~\ref{fig2}.

The LSTM and uRNN models are able to convincingly beat the baseline up to $T=400$ time steps.
Both models do well when $T=750$, but the mean squared error does not reach close to $0$.
The uRNN achieves lower test error, but it's curve is more noisy. Despite having vastly more parameters,
we monitored the LSTM performance to ensure no overfitting.

The RNN with tanh and IRNN were not able to beat the baseline for any number of time steps. 
\cite{Quoc2015} report that their RNN solve the problem for $T=150$ and the IRNN for $T=300$,
but they require over a million iterations before they start learning. 
Neither of the two models came close to either the uRNN or the LSTM in performance.
The stark difference in our findings are best explained by our use of RMSprop with significantly higher 
learning rates ($10^{-3}$ as opposed to $10^{-8}$) than \cite{Quoc2015} use for SGD with momentum.

\vspace{-0.25mm}
\subsection{Pixel-by-pixel MNIST}

In this task, suggested by \cite{Quoc2015}, algorithms are fed pixels of MNIST \citep{MNIST} sequentially and required to output a 
class label at the end. We consider two tasks: one where pixels are read in order (from left to right,
bottom to top) and one where the pixels are all randomly permuted using the same randomly generated
permutation matrix. The same model architectures as for the adding problem were used for this task, except we now use a 
softmax for category classification. We ran the optimization algorithms until convergence of the 
mean categorical cross entropy on test data, and plot test accuracy in Figure~\ref{fig3}.

Both the uRNN and LSTM perform applaudably well here.
On the correct unpermuted MNIST pixels, the LSTM performs better, achieving 98.2 \% 
test accurracy versus 95.1\% for the uRNN. 
However, when we permute the ordering of the pixels, the uRNN dominates with 91.4\% of accuracy in 
contrast to the 88\% of the LSTM, despite having less than a quarter of the parameters. 
This result is state of the art on this task, beating the IRNN \citep{Quoc2015}, which reaches 
close to 82\% after 1 million training iterations. Notice that uRNN reaches convergence in less 
than 20 thousand iterations, while it takes the LSTM from 5 to 10 times as many to finish learning.

Permuting the pixels of MNIST images creates many longer term dependencies across pixels than in the 
original pixel ordering, where a lot of structure is local. This makes it necessary for a network to 
learn and remember more complicated dependencies across varying time scales. 
The results suggest that the uRNN is better able to deal with such structure over the data, where the 
LSTM is better suited to more local sequence structure tasks.

\subsection{Exploratory experiments}

\begin{figure*}[t!] 
  \begin{minipage}[b]{0.25\linewidth}
    \centering
    \includegraphics[scale=0.15]{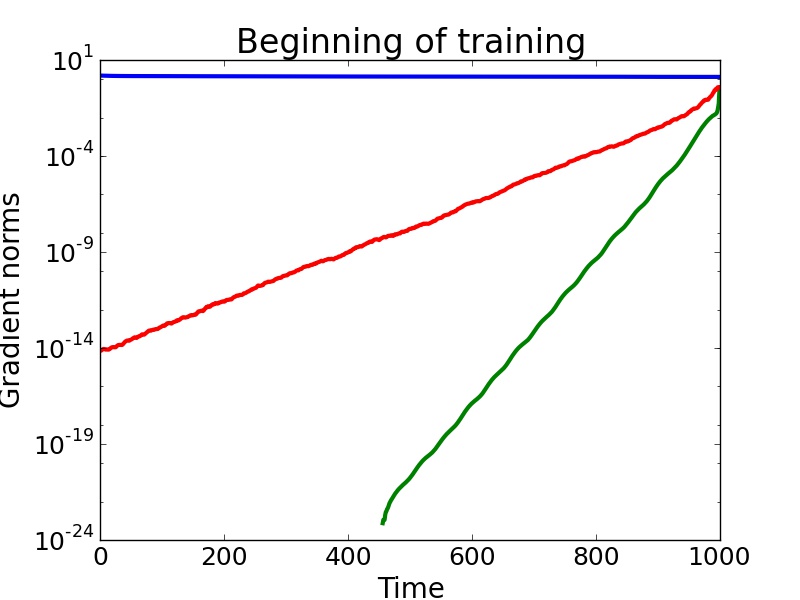}
  \end{minipage}
  \begin{minipage}[b]{0.25\linewidth}
    \centering
    \includegraphics[scale=0.15]{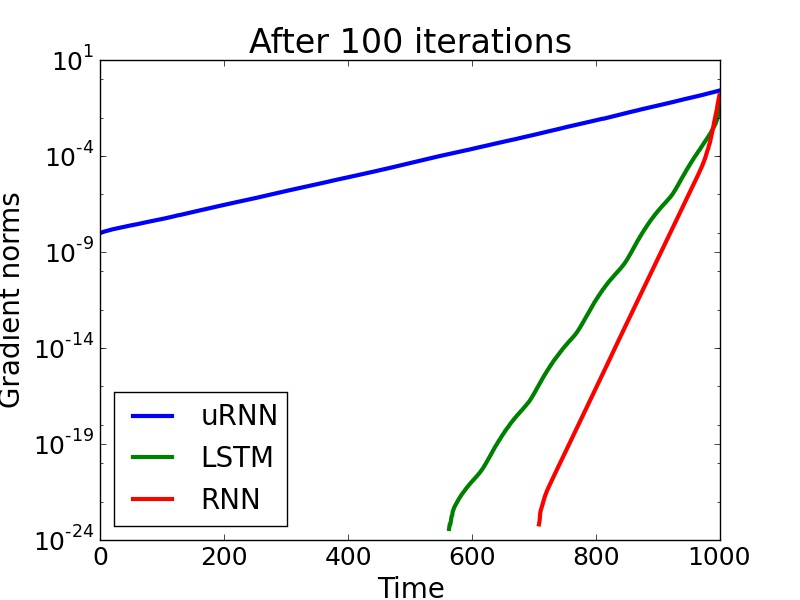}
  \end{minipage}
  \begin{minipage}[b]{0.25\linewidth}
    \centering
    \includegraphics[scale=0.15]{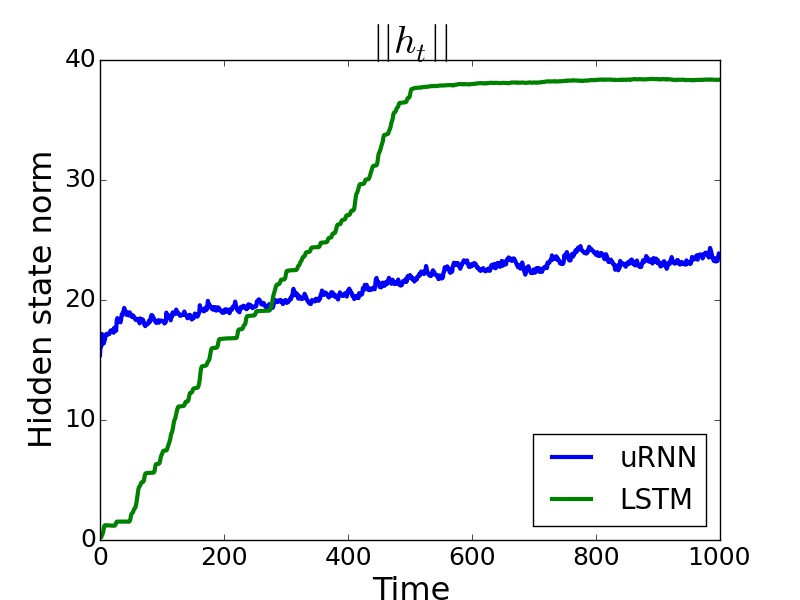}
  \end{minipage}
  \begin{minipage}[b]{0.25\linewidth}
    \centering
    \includegraphics[scale=0.15]{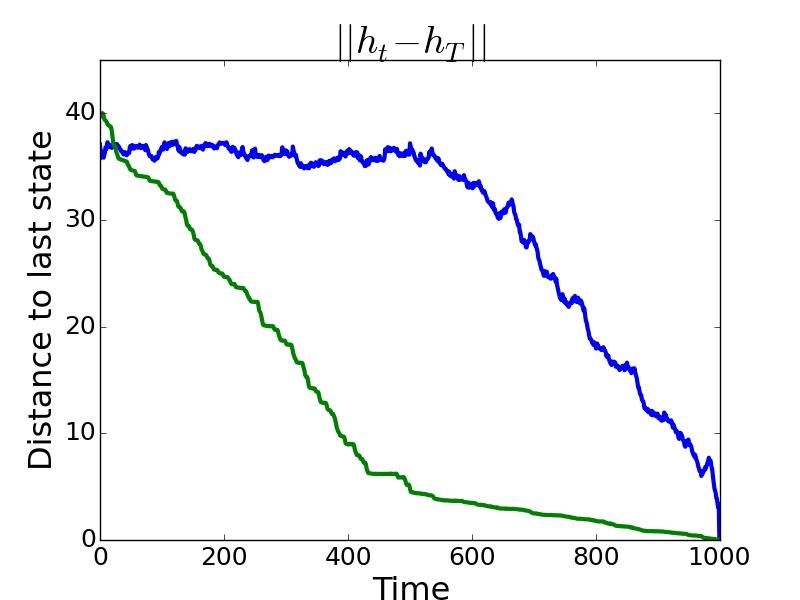}
  \end{minipage}
  \caption{From left to right. Norms of the gradients with respect to hidden states i.e.  
   $\norm{\frac{\partial C}{\partial h_t}}$ at (i) beginning of training, (ii) after 100 iterations.
   (iii) Norms of the hidden states and (iv) $L_2$ distance between hidden states and final hidden state.
   The gradient norms of uRNNs do not decay as fast as for other models as training progresses.
uRNN hidden state norms stay much more consistent over time than the LSTM. 
         LSTM hidden states stay almost the same after a number of time steps, 
         suggesting that it is not able to use new input information.
}
  \label{fig:gradnorms} 
\end{figure*}

{\bf Norms of hidden state gradients.} As discussed in Section~\ref{sec:ortho}, key to being able to learn long term 
dependencies is in controlling $\frac{\partial C}{\partial h_t}$. 
With this in mind, we explored how each model propagated gradients, by examining 
$\norm{\frac{\partial C}{\partial h_t}}$ as a function of $t$. 
Gradient norms were computed at the beginning of training and again after 100 iterations of
training on the adding problem. The curves are plotted in Figure~\ref{fig:gradnorms}.
It is clear that at first, the uRNN propagates gradients perfectly, while each other model has 
exponentially vanishing gradients. 
After 100 iterations of training, each model experiences vanishing gradients, 
but the uRNN is best able to propagate information, having much less decay.

\vspace{1mm}
{\bf Hidden state saturation.}
We claim that typical recurrent architectures saturate, in the 
sense that after they acquire some information, it becomes much more difficult to acquire further information pertaining
to longer dependencies.  
We took the uRNN and LSTM models trained on the adding problem with $T=200$, and computed 
a forward pass with newly generated data for the adding problem with $T=1000$. In order to show saturation
effects, we plot the norms of the hidden states and the $L_2$ distance between each state and the last in
Figure~\ref{fig:gradnorms}.    

In our experiments, it is clear that the uRNN does not suffer as much as other models do.
Notice that whilst the norms of hidden states in the uRNN grow very steadily over time, in the LSTM
they grow very fast, and then stay constant after about $500$ time steps. This behaviour may
suggest that the LSTM hidden states saturate in their ability to incorporate new information, which is vital
for modeling long complicated sequences. It is interesting to see that the LSTM hidden state at $t=500$, 
is close to that of $t=1000$, whilst this is far from the case in the uRNN. Again, this suggests that
the LSTM's capacity to use new information to alter its hidden state severly degrades with sequence length.
The uRNN does not suffer from this difficulty nearly as badly. 

A clear example of this phenomenon was observed in the adding problem with $T=750$. We found that the 
Pearson correlation between the LSTM output prediction and the first of the two uniform samples (whose 
sum is the target output) was $\rho = 0.991$. This suggests that the LSTM learnt to simply find and store the 
first sample, as it was unable to 
incorporate any more information by the time it reached the second, due to 
saturation of the hidden states.   

\vspace{-2.5mm}
\section{Discussion}
\vspace{-1mm}
There are a plethora of further ideas that may be explored from our findings, both with regards to 
learning representation and efficient implementation. For example, one hurdle of modeling long sequences 
with recurrent networks is the requirement of storing all hidden state values for the purpose of gradient
backpropagation. This can be prohibitive, since GPU memory is typically a limiting factor of neural network 
optimization. However, since our weight matrix is unitary, its inverse is its conjugate transpose, 
which is just as easy to operate with. If further we were to use an invertible nonlinearity function, we would 
no longer need to store hidden states, since they can be recomputed in the backward pass. This could have 
potentially huge implications, as we would be able to reduce memory usage by an order of $T$, the number of
time steps. This would make having immensely large hidden layers possible, perhaps enabling vast memory representations.

In this paper we demonstrate state of the art performance on hard problems requiring
long term reasoning and memory. These results are based on a novel parameterization of unitary matrices which permit
efficient matrix computations and parameter optimization. Whilst complex domain modeling has been 
widely succesful in the signal processing community (e.g. Fourier transforms, wavelets), we have yet to 
exploit the power of complex valued representation in the deep learning community. Our hope is that this
work will be a step forward in this direction. We motivate the idea of unitary evolution as a novel way
to mitigate the problems of vanishing and exploding gradients. Empirical evidence suggests that our uRNN is better able
to pass gradient information through long sequences and does not suffer from saturating hidden states as much
as LSTMs, typical RNNs, or RNNs initialized with the identity weight matrix (IRNNs).


\bigskip
{\bf Acknowledgments}: We thank the developers of
Theano~\citep{Fred2010} for 
their great work. We thank NSERC, Compute Canada, Canada Research Chairs and CIFAR for their support. We would also like to thank \c{C}aglar Gul\c{c}ehre, David Krueger, Soroush Mehri, Marcin Moczulski, Mohammad Pezeshki and Saizheng Zhang for helpful discussions, comments and code sharing.

\bibliography{uRNN}

\begin{thebibliography}{23}
\providecommand{\natexlab}[1]{#1}
\providecommand{\url}[1]{\texttt{#1}}
\expandafter\ifx\csname urlstyle\endcsname\relax
  \providecommand{\doi}[1]{doi: #1}\else
  \providecommand{\doi}{doi: \begingroup \urlstyle{rm}\Url}\fi

\bibitem[Bengio et~al.(1994)Bengio, Simard, and Frasconi]{Yoshua94}
Bengio, Yoshua, Simard, Patrice, and Frasconi, Paolo.
\newblock Learning long-term dependencies with gradient descent is difficult.
\newblock \emph{IEE Transactions on Neural Networks}, 5, 1994.

\bibitem[Bergstra et~al.(2010)Bergstra, Breuleux, Bastien, Lamblin, Pascanu,
  Desjardins, Turian, Warde-Farley, and Bengio]{Fred2010}
Bergstra, James, Breuleux, Olivier, Bastien, Fr{\'{e}}d{\'{e}}ric, Lamblin,
  Pascal, Pascanu, Razvan, Desjardins, Guillaume, Turian, Joseph, Warde-Farley,
  David, and Bengio, Yoshua.
\newblock Theano: a {CPU} and {GPU} math expression compiler.
\newblock \emph{Proceedings of the Python for Scientific Computing Conference
  ({SciPy})}, 2010.

\bibitem[Cho et~al.(2014)Cho, van Merri\"enboer, Bahdanau, and
  Bengio]{Cho2014a}
Cho, Kyunghyun, van Merri\"enboer, Bart, Bahdanau, Dzmitry, and Bengio, Yoshua.
\newblock On the properties of neural machine translation: {E}ncoder--{D}ecoder
  approaches.
\newblock In \emph{Eighth Workshop on Syntax, Semantics and Structure in
  Statistical Translation}, October 2014.

\bibitem[Collobert et~al.(2011)Collobert, Weston, Bottou, Karlen, Kavukcuoglu,
  and Kuksa]{Collobert2011}
Collobert, Ronan, Weston, Jason, Bottou, L{\'e}on, Karlen, Michael,
  Kavukcuoglu, Koray, and Kuksa, Pavel.
\newblock Natural language processing (almost) from scratch.
\newblock \emph{Journal of Machine Learning Research}, 12:\penalty0 2493--2537,
  2011.

\bibitem[Glorot \& Bengio(2010)Glorot and Bengio]{Glorotinit}
Glorot, Xavier and Bengio, Yoshua.
\newblock Understanding the difficulty of training deep feedforward neural
  networks.
\newblock \emph{International Conference on Artificial Intelligence and
  Statistics (AISTATS)}, 2010.

\bibitem[Glorot et~al.(2011)Glorot, Bordes, and Bengio]{Glorot2011}
Glorot, Xavier, Bordes, Antoine, and Bengio, Yoshua.
\newblock Deep sparse rectifier neural networks.
\newblock \emph{International Conference on Artificial Intelligence and
  Statistics (AISTATS)}, 2011.

\bibitem[Graves et~al.(2014)Graves, Wayne, and Danihelka]{NTM}
Graves, Alex, Wayne, Greg, and Danihelka, Ivo.
\newblock Neural turing machines.
\newblock \emph{arXiv preprint arXiv:1410.5401}, 2014.

\bibitem[Hinton et~al.(2012)Hinton, Deng, Yu, Dahl, Mohamed, Jaitly, Senior,
  Vanhoucke, Nguyen, Sainath, and Kingsbury]{Hinton2012}
Hinton, Geoffrey, Deng, Li, Yu, Dong, Dahl, George, Mohamed, Abdel-rahman,
  Jaitly, Navdeep, Senior, Andrew, Vanhoucke, Vincent, Nguyen, Patrick,
  Sainath, Tara, and Kingsbury, Brian.
\newblock Deep neural networks for acoustic modeling in speech recognition.
\newblock \emph{Signal Processing Magazine}, 2012.

\bibitem[Hirose(2003)]{hirose2003complex}
Hirose, Akira.
\newblock \emph{Complex-valued neural networks: theories and applications},
  volume~5.
\newblock World Scientific Publishing Company Incorporated, 2003.

\bibitem[Hochreiter(1991)]{Hochreiter91-small}
Hochreiter, S.
\newblock {Untersuchungen zu dynamischen neuronalen Netzen. Diploma thesis,
  T.U. M\"{u}nich}, 1991.

\bibitem[Hochreiter \& Schmidhuber(1997)Hochreiter and Schmidhuber]{LSTM}
Hochreiter, Sepp and Schmidhuber, J\"urgen.
\newblock Long short-term memory.
\newblock \emph{Neural Computation}, 8(9):\penalty0 1735--1780, 1997.

\bibitem[Hoffman \& Kunze(1971)Hoffman and Kunze]{linalgbook}
Hoffman, Kenneth and Kunze, Ray.
\newblock \emph{Linear Algebra}.
\newblock Pearson, second edition, 1971.

\bibitem[Krizhevsky et~al.(2012)Krizhevsky, Sutskever, and
  Hinton]{Krizhevsky2012}
Krizhevsky, Alex, Sutskever, Ilya, and Hinton, Geoffrey~E.
\newblock Imagenet classification with deep convolutional neural networks.
\newblock \emph{Neural Information Processing Systems}, 2012.

\bibitem[Le et~al.(2010)Le, Sarl\'os, and Smola]{fastfood}
Le, Quoc, Sarl\'os, Tam\'as, and Smola, Alex.
\newblock Fastfood - approximating kernel expansions in loglinear time.
\newblock \emph{International Conference on Machine Learning}, 2010.

\bibitem[Le et~al.(2015)Le, Navdeep, and Hinton]{Quoc2015}
Le, Quoc~V., Navdeep, Jaitly, and Hinton, Geoffrey~E.
\newblock A simple way to initialize recurrent networks of rectified linear
  units.
\newblock \emph{arXiv preprint arXiv:1504.00941}, 2015.

\bibitem[LeCun et~al.(1998)LeCun, Bottou, Bengio, and Haffner]{MNIST}
LeCun, Yann, Bottou, L\'eon, Bengio, Yoshua, and Haffner, Patrick.
\newblock Gradient-based learning applied to document recognition.
\newblock \emph{Proceedings of the IEEE}, 1998.

\bibitem[Martens \& Sutskever(2011)Martens and Sutskever]{HF}
Martens, James and Sutskever, Ilya.
\newblock Learning recurrent neural networks with hessian-free optimization.
\newblock \emph{International Conference on Machine Learning}, 2011.

\bibitem[Nair \& Hinton(2010)Nair and Hinton]{Nair2010}
Nair, Vinod and Hinton, Geoffrey~E.
\newblock Rectified linear units improve restricted boltzmann machines.
\newblock \emph{International Conference on Machine Learning}, 2010.

\bibitem[Pascanu et~al.(2010)Pascanu, Mikolov, and Bengio]{Pascanu2013}
Pascanu, Razvan, Mikolov, Tomas, and Bengio, Yoshua.
\newblock On the difficulty of training recurrent neural networks.
\newblock \emph{International Conference on Machine Learning}, 2010.

\bibitem[Saxe et~al.(2014)Saxe, McLelland, and Ganguli]{Saxe2014}
Saxe, Andrew~M., McLelland, James~L., and Ganguli, Surya.
\newblock Exact solutions to the nonlinear dynamics of learning in deep linear
  neural networks.
\newblock \emph{International Conference in Learning Representations}, 2014.

\bibitem[Tieleman \& Hinton(2012)Tieleman and Hinton]{RMSPROP}
Tieleman, Tijmen and Hinton, Geoffrey.
\newblock Lecture 6.5-rmsprop: Divide the gradient by a running average of its
  recent magnitude.
\newblock \emph{Coursera: Neural Networks for Machine Learning}, 2012.

\bibitem[Yang et~al.(2015)Yang, Moczulski, Denil, de~Freitas, Smola, Song, and
  Wang]{dfc}
Yang, Zichao, Moczulski, Marcin, Denil, Misha, de~Freitas, Nando, Smola, Alex,
  Song, Le, and Wang, Ziyu.
\newblock Deep fried convnets.
\newblock \emph{International Conference on Computer Vision (ICCV)}, 2015.

\bibitem[Zimmermann et~al.(2011)Zimmermann, Minin, and
  Kusherbaeva]{zimmermann2011comparison}
Zimmermann, Hans-Georg, Minin, Alexey, and Kusherbaeva, Victoria.
\newblock Comparison of the complex valued and real valued neural networks
  trained with gradient descent and random search algorithms.
\newblock In \emph{ESANN}, 2011.

\end{thebibliography}
\bibliographystyle{icml2016}

\end{document}